# End-to-end Graph Learning Approach for Cognitive Diagnosis of Student Tutorial

Fulai Yang
*College of Computer Science and Technology*
*Chongqing University of Posts and Telecommunications*
Chongqing, China
ab3354349977@gmail.com

Di Wu
*College of Computer and Information Science*
*Southwest University*
Chongqing, China
wudi.cigit@gmail.com

Yi He
*School of Data Science*
*William & Mary*
Williamsburg, USA
yihe@wm.edu

Li Tao
*College of Computer and Information Science*
*Southwest University*
Chongqing, China
tli@swu.edu.cn

Xin Luo
*College of Computer and Information Science*
*Southwest University*
Chongqing, China
luoxin@swu.edu.cn

*Abstract*—Cognitive diagnosis (CD) utilizes students' existing studying records to estimate their mastery of unknown knowledge concepts, which is vital for evaluating their learning abilities. Accurate CD is extremely challenging because CD is associated with complex relationships and mechanisms among students, knowledge concepts, studying records, etc. However, existing approaches loosely consider these relationships and mechanisms by a non-end-to-end learning framework, resulting in sub-optimal feature extractions and fusions for CD. Different from them, this paper innovatively proposes an End-to-end Graph Neural Networks-based Cognitive Diagnosis (EGNN-CD) model. EGNN-CD consists of three main parts: knowledge concept network (KCN), graph neural networks-based feature extraction (GNNFE), and cognitive ability prediction (CAP). First, KCN constructs CD-related interaction by comprehensively extracting physical information from students, exercises, and knowledge concepts. Second, a four-channel GNNFE is designed to extract high-order and individual features from the constructed KCN. Finally, CAP employs a multi-layer perceptron to fuse the extracted features to predict students' learning abilities in an end-to-end learning way. With such designs, the feature extractions and fusions are guaranteed to be comprehensive and optimal for CD. Extensive experiments on three real datasets demonstrate that our EGNN-CD achieves significantly higher accuracy than state-of-the-art models in CD.

## I. Introduction

Cognitive Diagnosis(CD) is the use of students' existing learning records to assess their mastery of unknown knowledge concepts[1]. CD is a multidisciplinary field[2-8] at the intersection of cognitive psychology, educational assessment, and psychometrics[9-13]. It aims to understand and diagnose the underlying cognitive processes of human learning and problem solving [14][25] which is relevant to our lives. Specifically, in the intelligent education system[15-18], CD aims at a comprehensive assessment of students' individual abilities. Fig. 1 shows a toy example of CD. In general, the teacher will select several exercises (e.g., $e_1…e_5$) from the exercise bank to form a test paper, and the students will answer the corresponding exercises (e.g., right or wrong). Given student studying records, Our purpose is to reveal the specific cognitive skills(e.g., *Infinite series*) and knowledge structures possessed by individuals[19], thereby accurately assessing their personal abilities[20-24].

CD has been developed for decades and can be divided into two main categories. The first is the CD based on statistical methods such as DINA[37], IRT[38], IRR-IRT[29]. The other is CD based on simple Graph Neural Network(GNN), such as NeuralCDM[30], KaNCD[30] and TechCD[32]. These models are all representative models in CD. However, these models adopt linear principles or some non-end-to-end neural networks to explain the complex process of students solving problems[26-28]. It is far from enough to effectively capture the potential personal information of individual students, but it is necessary to extract student information from a global perspective[33-35]. So it is impossible to model students' learning behavior in complex educational platforms[36][37], and it also fails to perform well in the face of large-scale datasets[38-43].

Existing approaches loosely consider these relationships and mechanisms by a non-end-to-end learning framework, resulting in sub-optimal feature extractions and fusions for CD. In order to extract feature information from studying records more effectively, this paper proposes an End-to-end Graph Neural Networks-based Cognitive Diagnosis (EGNN-CD) model[44][48][49]. EGNN-CD consists of three main parts: knowledge concept network (KCN), GNN feature extraction (GNNFE) and cognitive ability prediction (CAP). First, KCN is an infographic that syntheses physical information about students, exercises, and knowledge concepts to build CD-related interactions. Second, GNNFE designed a four-channel graph neural network to learn the higher-order features and individual features of KCN end-to-end. It ensuring that feature extraction and subsequent fusion of CD events are comprehensive and optimal predictions. Finally, CAP combines high-level information to predict a student's ability. With such designs, the feature extractions and fusions are guaranteed to be comprehensive and optimal for CD. Experiments on three real datasets proved from all aspects that EGNN-CD has higher accuracy and significance in CD, which is superior to the existing model.

The contributions of this article mainly include the following aspects:

- This paper proposes an end-to-end cognitive diagnosis model based on graph neural network, which solves the CD problem effectively.

- A tripartite knowledge network of student-exercise-knowledge concept is constructed to convey higher-order information.
- A novel four-channel end-to-end graph neural network constructed to efficiently learn high-level information of students and exercises from a global perspective.

Extensive experiments on three real datasets have illustrated the superior performance of EGNN-CD model in CD from various aspects.

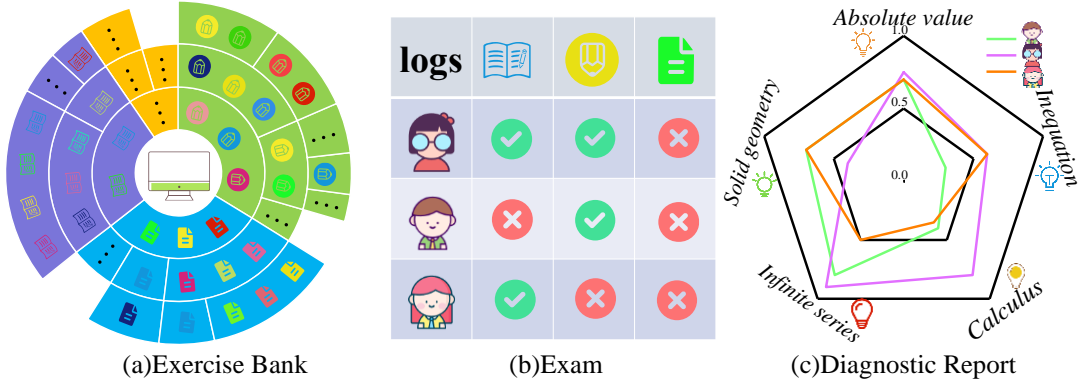

(a)Exercise Bank    (b)Exam    (c)Diagnostic Report

Fig. 1. A toy example of cognitive diagnosis

## II. RELATED WORK

CD is a multidisciplinary field that intersects cognitive psychology, educational assessment and psychometrics. Its main purpose is to simulate the potential process of human learning and solving new problems. CD models can be roughly divided into the following two categories.

DINA[37] and IRT[38] are the most classic CD models. The DINA[37] model is a discrete CD model. This model exploits a knowledge concept mastery vector to model students' mastery of knowledge. The impact of HO-DINA low-latitude potential characteristics on students[39]. FuzzyCDF [40] fuzzy the examinee's skill proficiency, combined with fuzzy set theory and educational assumptions, established a problem mastery model based on the examinee's skill proficiency. IRT [38] is a typical continuous CD model. IRT jointly models the test questions and students through the students' ($\theta$) answering conditions. thereby deriving test question parameters and students' potential abilities. Based on IRT, MIRT [31] transforms the one-dimensional latent features of students ($\theta$) into multi-dimensional latent features $\theta_i=c(\theta_{i1},\theta_{i2},…, \theta_{ik})$, where $k$ is the dimension referred to in MIRT. IRR-IRT [29]introduces pairwise learning and transforms it into CD to model the monotonicity between item responses. These models rely too much on artificially designed estimation functions and handle the differences between subjects with different ability levels in tests, but can only model one knowledge point of a student and cannot reflect the relationship between students and knowledge concepts [51][52]. Thus their oversimplification of cognitive processes leads to limited fitting ability and difficulty in handling enormous datasets [45-47].

In recent years, scholars have continuously integrated neural networks into CD and achieved some remarkable results. EIRS[49] is a general framework that explores the partial ordering between interacted and uninteracted exercises. Wang et al [50] integrated the explainability of CD to educational priors into a deep learning-based knowledge tracking method and proposed dynamic CD. Li et al[53] applied CD computer adaptive testing to CD to improve classification accuracy. Ma et al [54] utilize minimal model assumptions to jointly learn the problem of these latent hierarchical structures in CD from observational data. The most representative method is NeuralCDM [30]. It plans students and exercises to consider factor vectors and combines neural networks to simulate complex learning processes. KaNCD[30] considers the associated information between knowledge concepts based on NeuralCDM. TechCD[32] operates a graph convolutional network with bottom discard operations to learn the characteristics of students and exercises, allowing for cross-domain learning[55][58]. Although these models can achieve good results by modeling each module separately, they do not comprehensively consider the mutual influence and are prone to fall into local optimality. At the same time, the feature extraction and fusion results of these models are not back-transferred, so the problem cannot be modeled from a global perspective[56-63].

## III. PRELIMINARY

### A. Notations and Symbols

The symbols used in the article and their explanations are shown in the Table I.

### B. Data Definition

a. Student-Exercise Matrix. The Student-Exercise Matrix represents the students' answers in the question bank. It is represented by $y \in (0, y_{nm})^{N \times M}$, where $N$ and $M$ represent the number of students and the number of exercises in the question bank, respectively. $y$ represents the corresponding score. That is, $y=y_{nm}$ means that student $s_n$ does exercise $e_m$ and gets a score of $y_{nm}$, and if $y=0$, it means that student $s_n$ doesn't exercise $e_m$.

b. Student-Knowledge Concept matrix. The Student - Knowledge Concept Matrix indicates the students' mastery of knowledge concepts during the learning process. It is expressed as $y_{nc} \in (0,1)^{N \times C}$, where $N$ and $C$ represent the number of students and the number of knowledge concepts in the question bank, respectively. $y_{nc}$ represents the specific mastery situation, that is, $y_{nc}=1$ means that student $s_n$ has mastered knowledge concept $k_c$, and if $y_{nc}=0$, it means that student $s_n$ has not mastered knowledge concept $k_c$.

c. Exercise-Knowledge Concept matrix. The Exercise- Knowledge Concept matrix represents the inclusion relationship between exercises and knowledge concepts. It is usually called the Q-matrix, which is manually labeled by education platform experts and is represented as $Q_{mc}=(0,1)^{M \times C}$. $Q_{mc}=1$ which represents the exercise question $e_m$ contains the knowledge concept $k_c$, otherwise $Q_{mc}=0$.

TABLE I. NOTATIONS AND EXPLANATIONS

| Notation | Explanation |
|---|---|
| $S, s_n$ | Set of student and an student entity |
| $E, e_m$ | Set of exercise and an exercise entity |
| $K, k_c$ | Set of knowledge concept and an knowledge concept entity |
| $R_{(S,E,Y)}$ | Set of studying records |
| $log(s_n, e_m, y_{nm})$ | A studying record of $m$-th exercise done by $n$-th student |
| $Q$ | The relationship matrix between exercises and knowledge concepts |
| $d$ | Output dimensions of the GNN module |
| $l_{ij}$ | Layer $i$ of the $GNN_j$ module |
| $\|\cdot\|_2$ | The L2 norm of a vector |
| $\|\cdot\|$ | If the element is a set, it means to take the number of elements in the set. |

## C. Problem Definition

Suppose there are $N$ students, $M$ exercises, and $C$ knowledge concepts in a system, and they can be expressed as $S=\{s_1,s_2,...,s_N\}, E=\{e_1,e_2,...,e_M\}$ and $K=\{k_1,k_2,...,k_C\}$. Each student has done some exercise questions in the question bank. The studying records $R_{(S,E,Y)}$ represents the record of all students doing the questions. Each record $log(s_n, e_m, y_{nm})$ represents the score $y_{nm}$ that the student $s_n$ received after doing the exercise question $e_m$. Given studying records $R_{(S,E,Y)}$, the purpose of CD is to learn the student's problem-solving process to model the student's ability. i.e.

$$\mathbb{F}_{CDM}(s,e,\Theta) \to y_{nm} \qquad (1)$$

where $s$ and $e$ are feature representation vectors for students and exercises. $F_{CDM}(\cdot)$ is the CD function. $\Theta$ is the parameter required for training. And $\hat{y}_{nm}$ is the predicted performance score.

## IV. PROPOSED MODEL

The EGNN-CD model mainly contains three modules, namely knowledge concept network (KCN), GNN feature extraction (GNNFE) and cognitive ability prediction (CAP).

### A. Knowledge concept network

The original studying records solely contains the student id/exercise id and the corresponding score. If we directly exploit this information into the GNN module training. We will merely get two parts of information, and this information is the fusion of student and exercise information. It cannot clearly express the students' learning ability and the characteristics of the exercises, and it is not conducive to subsequent result predictions. To this end, high-level information needs to be extracted. The specific form of the initial information is shown in (2). We thus build a tripartite feature knowledge network of students-exercises-knowledge concepts to convey high-level information[32][31]. At the same time, one-hot encoding is used to distinguish each entity. So that the unique high-order information of each entity will be transmitted along KCN during end-to-end learning.

$$s_n = [\underbrace{e_1^{(0)}, e_2^{(0)}, ..., e_M^{(0)}}_{exercise-feature}, \underbrace{k_1^{(0)}, k_2^{(0)}, ..., k_C^{(0)}}_{concept-feature}],$$

$$e_m = [\underbrace{s_1^{(0)}, s_2^{(0)}, ..., s_N^{(0)}}_{student-feature}, \underbrace{k_1^{(0)}, k_2^{(0)}, ..., k_C^{(0)}}_{concept-feature}] \qquad (2)$$

### B. GNN Feature Extraction

Most existing models ply linear or simple neural networks to learn students' learning processes. Although amazing results have been achieved, existing approaches loosely consider these relationships and mechanisms by a non-end-to-end learning framework. It resulting in sub-optimal feature extractions and fusions for CD. And the Q-matrix is given by relevant expert marks, and there are personal subjective factors of experts. GNNFE has subsequently design a multi-layer perceptron. It ensures that feature extraction and subsequent fusion of CD events are comprehensive and optimal predictions through end-to-end learning of pre-fusion student and exercise features. As mentioned above, the information extraction process should be conducted independently to reduce negative interactions between the student and the exercise. For this purpose, we design a 4-channel GNN module to learn high-level information of students and exercises in KCN end-to-end, as follows:

$$s_n = \Phi_s[GNN_1(x_n^{se}), GNN_2(x_n^{sk})] \quad (3)$$

$$e_m = \Phi_e[GNN_3(x_m^{es}), GNN_4(x_m^{ek})] \quad (4)$$

$x_n^{se} \in (0,1)^{1 \times M}$ and $x_n^{sk} \in (0,1)^{1 \times C}$ represent the information of the *n*-th student's mastery of exercises and knowledge concept respectively. $x_m^{es} \in (0,1)^{1 \times N}$ and $x_m^{ek} \in (0,1)^{1 \times C}$ represent the *m*-th exercise's property and the relationship between *m*-th exercise and knowledge concept respectively. $\Phi_s$ and $\Phi_e$ represent information aggregation operations. As shown in the Fig. 2, when processing each feature map, we can select multiple GNN modules. After that perform a weighted summation of the information processed by these modules, it is formulated as follow. $x$ stands for $x_n^{se}$, $x_n^{sk}$, $x_m^{es}$ and $x_m^{ek}$ in that order. $x_{mid}$ represents the corresponding output and takes $d=128$. $l$ represents the number of layers of GNN. $\sigma$ is the activation function. $W_i^{(g)}$ and $b_i^{(g)}$ are trainable parameters. Moreover, *dropout* strategy is adopted to prevent the training process from overfitting.

## C. Cognitive Ability Prediction

CAP eventually incorporates high-level information to make predictions about student ability. In order to maximize the information of the four paths, we consider their complementarity and correlation in the fused end-to-end neural network layer. For student $s_n$ and exercise $e_m$, we can obtain information representations from the graph neural network, represented by $s_{n1}$, $s_{n2}$, $e_{m1}$ and $e_{m2}$ respectively. The fusion operation can be described by the following formula: $\hat{y}_{nm} \in (0,1)^{1 \times 1}$ represents the final prediction result. $\sigma$ is the activation function. $\Phi(\bullet)$ indicates the information aggregation function, where direct concatenation is used. $W^{(P)}$ and $b^{(P)}$ are trainable parameters. In addition, the student learning process is progressive [38]. That is, as the number of effective questions increases, the student's ability will gradually improve. Therefore, in the training process to ensure $W^{(p)} \geq 0$. *dropout* strategy is adopted to prevent overfitting during the training process. Besides, the cross-entropy function is exerted as the loss function during training.

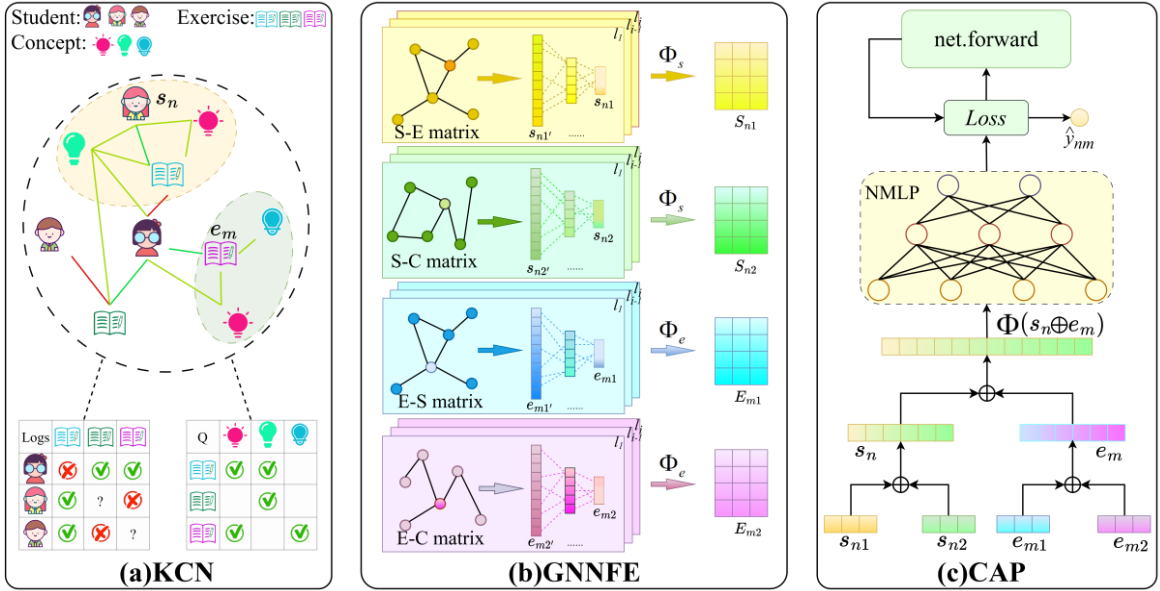

Fig. 2. Structure of EGNN-CD, (a) Knowledge concept network(KCN), (b)GNN feature extraction(GNNFE) and (c)Cognitive ability prediction(CAP).

$$x_{mid} = \sum_{i=1}^{l} \frac{\exp(\|x\|_2)}{\sum_{i=1}^{l} \exp(\|x\|_2)} \times \sigma(W_i^{(g)} \times x + b_i^{(g)}), x_{mid} \in (0,1)^{1 \times d} \quad (5)$$

$$s_n = \Phi_s(s_{n1} \oplus s_{n2}), e_m = \Phi_e(e_{m1} \oplus e_{m2}) \quad (6)$$

$$y_{nm} = \sigma(W^P \times \Phi(s_n \oplus e_m) + b^P), W^P = Relu(W^P) \quad (7)$$

$$Loss = \sum_{R_{(S,E,Y)}} y_{nm} \log(\hat{y}_{nm}) + (1 - y_{nm}) \log(1 - \hat{y}_{nm}) \quad (8)$$

## D. Time Complexity

Based on the above analysis process, the EGNN-CD is proposed. Table II lists the execution complexity of each step of the model. Specifically, in general, $|S| \geq |E| \geq |K|$, the time complexity of the data fusion part is $\Theta(|S| \times |E|)$. The time complexity of the GNN processing part is $\Theta(|S| \times |E| \times d \times T)$. And each layer of the result prediction part contains a constant number of nodes, so its time complexity is approximately $\Theta(d \times T)$. Among them, d and T are constants, so the time complexity of the EGNN-CD is $\Theta(|S| \times |E|)$.

V. EXPERIMENT

In the following experiments, we mainly answer the following research questions:

- RQ.1. How does the proposed EGNN-CD model perform in terms of CD compared with existing models?
- RQ.2. The influence of each module of the EGNN-CD on the model performance?
- RQ.3. The influence of each hyperparameter on the performance of EGNN-CD?

*A. General Setting*

*a. Datasets*

Our experiments were conducted on three datasets, namely Junyi, Math1[40], Math2[40]. Table III summarizes the information of these datasets. 5-fold cross validation is employed. $AVG_{\#log}$ represents the average number of interactive knowledge concepts for each student. The specific method is shown in (10).

$$AVG_{\#log} = \frac{\sum_n^S \sum_c^K I(log(s_n, k_c, y_{nc}))}{|S|/|K|} \quad (9)$$

where $I(\cdot)=1$ if $log(s_n, k_c, y_{nc})$ exists, otherwise $I(\cdot)=0$. According to the data in Table III, Junyi contains 10,000 students, involving 835 knowledge points, but the number of knowledge points interacted by each student is 0.04, which is a large sparse dataset. Math1 and Math2 contain about 4,000 students, involving 11-16 knowledge points, but the number of knowledge points interacted by each student is 4.0-6.1, which is a small dense dataset. These three datasets are commonly used datasets for cognitive diagnosis problems and are very representative.

*b. Baseline*

Comparative models include discrete type DINA[37], continuous type IRT [38], MIRT [31], IRR-IRT [29], PMF using matrix decomposition, and the representative model TechCD[32], NeuralCDM [30] in recent years and its variant KaNCD [30]. These are the classic models of CD.

*c. Metric*

To demonstrate the performance of the EGNN-CD model in predicting student abilities, we employ 6 widely used metrics[64-66]: Accuracy (Acc), Area Under the Precision-Recall curve (AUPR), Area Under the Curve (AUC), F1 score(F1), Precision (Pre), Recall (Rec). Acc represents the probability of correct prediction for all samples. The higher the value, the better. If $y_{nm}$, $\hat{y}_{nm}$ are both greater than 0.5 or less than or equal to 0.5, then $f(y_{nm}, \hat{y}_{nm})=1$, otherwise $f(y_{nm}, \hat{y}_{nm})=0$. AUC and AUPR are designed to deal with the uneven division of positive and negative samples.

$$ACC = \frac{1}{\sum_{i=1}^N |E_i^{(Y)}|} \sum_{i=1}^N \sum_{e_j \in E_i^{(Y)}} f(y_{ij}, y_{ij}) \quad (10)$$

$$AUC = \int h(u)du, \quad AUPR = \int g(t)dt \quad (11)$$

Their values are between 0.5 and 1. The larger the value, the better the model's performance. Where $h(u)$ represents the ROC curve and $g(t)$ represents the PR curve. Pre and Rec are two indicators applied in statistics to measure the accuracy of a two-classification model and F1 takes into account both the precision and recall of the classification model. Its value is between 0 and 1. The larger the value, the better the model.

$$Pre = \frac{TP}{TP+FP}, Rec = \frac{TP}{TP+FN} \quad (12)$$

$$F1 = \frac{2 \times Pre \times Rec}{Pre + Rec} \quad (13)$$

Where True Positives (TP) represents the number of positive classes predicted as positive classes. False Positives (FP) represents the number of negative classes predicted as positive classes. and False Negatives (FN) represents the number of positive classes predicted as negative classes. True Negatives (TN) represents the number of negative classes predicted as negative classes.

*d. Training Setting*

In the experiment, we divide the dataset into the training set and the test set at a ratio of 8:2, and apply a 5-fold cross-validation experimental method to avoid the contingency of the results. Diverse datasets have unique training hyperparameter settings. We take the Junyi dataset as an example. EGNN-CD model learning rate of the comparison model is set to 0.003. The number of training rounds is 200. The parameters utilized in network training are initialized using the Xavier method, and the training optimizer adopts Adam[67][68].

TABLE II. COMPARISON OF FORECAST ACCURARY, INCLUDING WIN/TIE/LOSS, WILCOXON SIGNED-RANK TEST, AND FRIEDMAN TEST

| Dataset | Metric | EGNN-CD | KaNCD | IRT | PMF | NeuralCDM | IRR-IRT | MIRT | TechCD | DINA |
|---|---|---|---|---|---|---|---|---|---|---|
| D1 | Acc | **0.7802** | 0.7646 | 0.7689 | 0.7511 | 0.7465 | 0.7154 | 0.6983 | 0.7376 | 0.5016 |
| | AUPR | **0.8894** | 0.8842 | 0.8855 | 0.8737 | 0.8690 | 0.8519 | 0.8278 | 0.8219 | 0.7824 |
| | AUC | **0.8302** | 0.8170 | 0.8189 | 0.8015 | 0.7925 | 0.7646 | 0.7172 | 0.7470 | 0.6357 |
| | F1 | **0.8406** | 0.8283 | 0.8322 | 0.8152 | 0.8122 | 0.8067 | 0.7809 | 0.8120 | 0.5677 |
| | Pre | **0.7966** | 0.7895 | 0.7897 | 0.7892 | 0.7847 | 0.7235 | 0.7411 | 0.7625 | 0.6528 |
| | Rec | 0.8899 | 0.8716 | 0.8796 | 0.8431 | 0.8418 | 0.9115● | 0.8252 | 0.8690 | 0.5023 |
| D2 | Acc | **0.7511** | 0.7218 | 0.7311 | 0.7277 | 0.7259 | 0.7311 | 0.7261 | 0.7170 | 0.5522 |
| | AUPR | **0.8338** | 0.7735 | 0.7716 | 0.7679 | 0.7670 | 0.7710 | 0.7656 | 0.7446 | 0.6717 |
| | AUC | **0.8489** | 0.8152 | 0.8052 | 0.8097 | 0.8086 | 0.8045 | 0.8093 | 0.7885 | 0.7424 |
| | F1 | **0.7592** | 0.7237 | 0.7240 | 0.7247 | 0.7265 | 0.7230 | 0.7195 | 0.7372 | 0.3082 |
| | Pre | 0.7039 | 0.6899 | 0.708● | 0.6986 | 0.6920 | 0.7096● | 0.7022 | 0.6611 | 0.5832 |
| | Rec | **0.8255** | 0.7782 | 0.7408 | 0.7530 | 0.7648 | 0.7370 | 0.7378 | 0.8330 | 0.2095 |
| D3 | Acc | **0.8003** | 0.7222 | 0.7226 | 0.7209 | 0.7093 | 0.6781 | 0.7161 | 0.6984 | 0.5781 |
| | AUPR | **0.8680** | 0.7451 | 0.7390 | 0.7373 | 0.7229 | 0.6702 | 0.7278 | 0.7071 | 0.6337 |
| | AUC | **0.8885** | 0.8015 | 0.7916 | 0.7937 | 0.7818 | 0.7413 | 0.7881 | 0.7660 | 0.7223 |
| | F1 | **0.7771** | 0.6632 | 0.6737 | 0.6732 | 0.6651 | 0.6524 | 0.6719 | 0.6560 | 0.3305 |
| | Pre | **0.7581** | 0.7069 | 0.6939 | 0.6902 | 0.6707 | 0.6185 | 0.6795 | 0.6548 | 0.5407 |
| | Rec | **0.7987** | 0.6297 | 0.6547 | 0.6572 | 0.6601 | 0.6908 | 0.6647 | 0.6577 | 0.2381 |
| Statistic | Win/Tie/Loss | **141/0/3*** | 18/0/0 | 17/0/1 | 18/0/0 | 18/0/0 | 16/0/2 | 18/0/0 | 18/0/0 | 18/0/0 |
| | $p$-value | - | $1.07 \times 10^{-4}$ | $1.50 \times 10^{-4}$ | $1.07 \times 10^{-4}$ | $1.07 \times 10^{-4}$ | $2.10 \times 10^{-4}$ | $1.07 \times 10^{-4}$ | $1.27 \times 10^{-4}$ | $1.07 \times 10^{-4}$ |
| | F-rank | **1.22** | 3.83 | 3.25 | 4.22 | 5.28 | 5.92 | 6.00 | 6.28 | 9.00 |

## B. Performance Comparison(RQ.1)

The experimental results of the EGNN-CD model and the comparison model on diverse datasets is recorded in Table V. In order to better analyze the experimental results, we make statistical analyses of the Win/Tie/Loss[69], the Wilcoxon signed‑ranks test [70], and the Friedman test[71-74]. From Table V, we have the following significant observations:

- The EGNN-CD model proposed is excellent on the involved three datasets, and its total Win/Tie/Loss cases situation is 141/0/3 when comparing with each compared model one by one. Moreover, KaNCD and IRT, as representative models, also show wonderful results. While DINA performs poorly, this may be due to the discrete nature of the model itself.

- All p-values are much smaller than 0.05, indicating that the prediction accuracy of the EGNN-CD model is significantly higher than all the compared models.

- EGNN-CD model achieves the lowest F‑rank value among all the models, further confirming its highest prediction accuracy on the tested datasets.

TABLE III. EGNN-CD MODEL EXECUTION PROCESS

| Steps | Input: ($s_n$, $e_m$, $y_{nm}$) Output: $\hat{y}_{nm}$ | Cost |
|---|---|---|
| 1 | set $\lambda$=0.003, d=128, T=200 | $\Theta(1)$ |
| 2 | extract 4 feature maps from fusion information, respectively | $\Theta(|S| \times |E|)$ |
| 3 | **while** $t \leq T$ && not converge | $\Theta(T)$ |
| 4 | **for** $i$=1 to $|S|$ | $\times |S|$ |
| 5 | calculate $x_n^{se}$, $x_n^{sk}$ according to formula (5) | $\times |E| \times d$ |
| 6 | **end for** | - |
| 7 | **for** $j$=1 to $|E|$ | $\times |E|$ |
| 8 | calculate $x_m^{es}$, $x_m^{ek}$ according to formula (5) | $\times |S| \times d$ |
| 9 | **end for** | - |
| 10 | calculate $s_n$ and $e_m$ according to formula (6) | $\times \Theta(1)$ |
| 11 | calculate $\hat{y}_{nm}$ by formula (7),(8) | $\times \Theta(d)$ |
| 12 | $t=t+1$ | $\times \Theta(1)$ |
| 13 | **end while** | - |

TABLE IV. PROPERTIES OF THE THREE DATASETS

| No. | Name | $|S|$ | $|E|$ | $|K|$ | Response logs | concept per exercise | AVG$_{\#log}$ |
|---|---|---|---|---|---|---|---|
| D1 | Junyi | 10000 | 835 | 835 | 353835 | 1.0 | 0.04 |
| D2 | Math1 | 4209 | 20 | 11 | 84180 | 3.35 | 6.1 |
| D3 | Math2 | 3911 | 20 | 16 | 78220 | 3.2 | 4.0 |

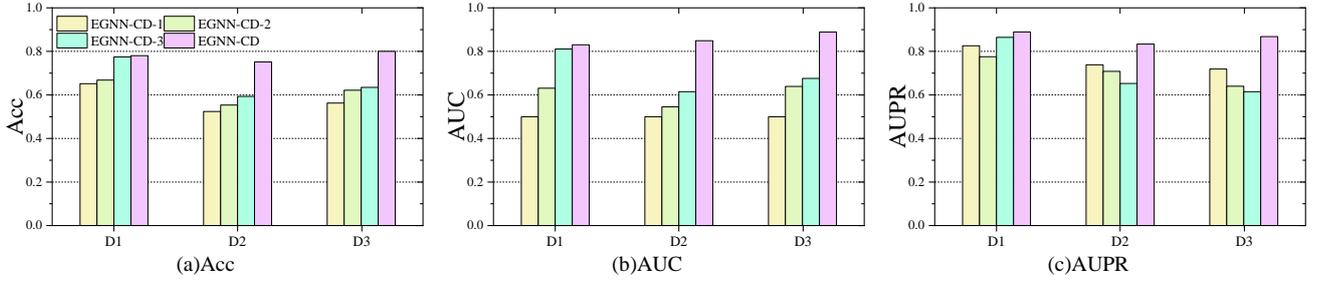

Fig. 3. Results of ablation experiment

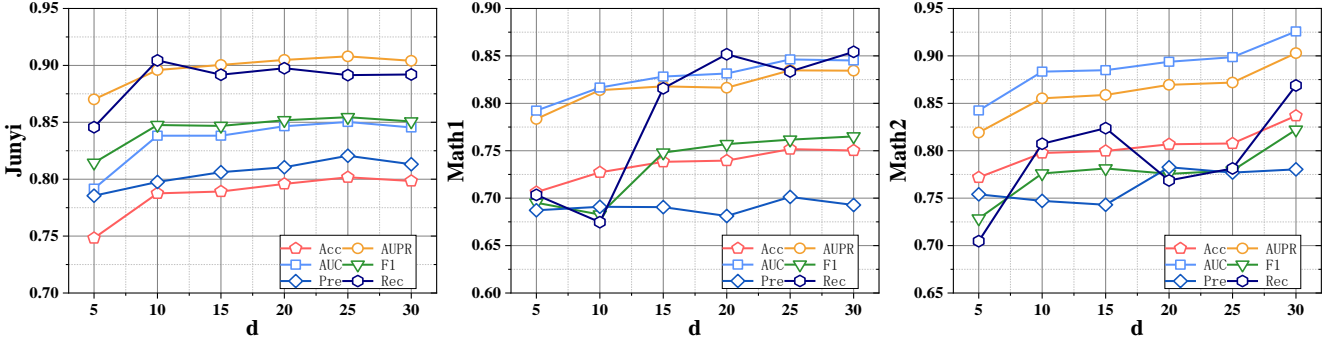

Fig. 4. The influence of dimension *d* on experimental results

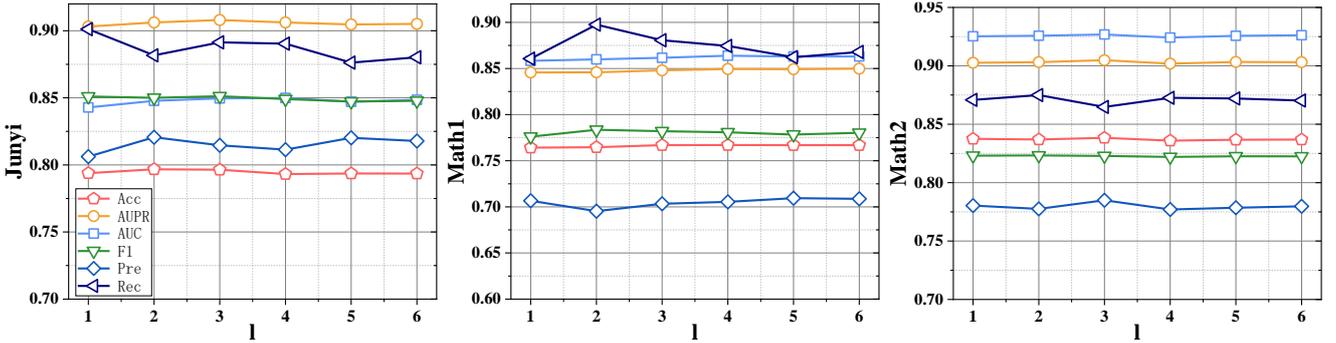

Fig. 5. The number of GNN layers influences the performance of the mode

TABLE V. VARIANTS OF EGNN-CD MODEL

| Model | Description |
|---|---|
| EGNN-CD -1 | Contains 1 GNN module, i.e. $GNN_1(x_n^{se})$ |
| EGNN-CD -2 | Contains 2 GNN modules, i.e. $GNN_1(x_n^{se})$ $GNN_2(x_n^{sk})$ |
| EGNN-CD -3 | Contains 3 GNN modules, i.e. $GNN_1(x_n^{se})$ $GNN_2(x_n^{sk})$ $GNN_3(x_m^{ex})$ |
| EGNN-CD | Contains 4 GNN modules, i.e. $GNN_1(x_n^{se})$ $GNN_2(x_n^{sk})$ $GNN_3(x_m^{ex})$ $GNN_4(x_m^{ek})$ |

Therefore, these observations demonstrate that EGNN-CD model significantly outperforms the comparison models in terms of accuracy in CD problem.

*C. Ablation Experiment(RQ.2)*

In order to better observe the impact of the GNN module on model performance, we design three variants of the EGNN-CD model. The structure of each model is shown in the Table IV. We apply each variant to each dataset and utilize a 5-fold cross-validation method to avoid the chance of experimental results. The experimental results are as shown in the Fig. 3. We draw some critical conclusions from the results in Fig. 3: 1) Fig. 3 illustrates the effectiveness of our model design. 2) as the number of EGNN-CD modules increases, the performance of each variant gradually increases in terms of Acc, AUC. But in terms of AUPR this is not the case. In Fig. 3 (c), the AUPR values of the D2 and D3 datasets first decrease and then increase. 3) we find that the AUC values of the EGNN-CD-1 variants were all 0.5, and the prediction results at this time were meaningless. This shows that using barely one EGNN-CD module cannot effectively model the student learning process.

*D. Hyperparameter Sensitivity Experiment(RQ.3)*

We conduct experiments on the influence of GNN output dimension on learning process modeling. From Fig. 4, we can see that with the increase of dimension $d$, although the performance of EGNN-CD model fluctuates somewhat, the overall trend is upward. It indicates that the more information in the modeling learning process, the more representative of the characteristics of students and exercises. When $d$=10, Pre and Rec fluctuate greatly in Math1 dataset, and their expression ability is one-sided. F1 value is a combination of Pre and Rec, and its value changes little, which indicates that F1 can better reflect the change of model performance. As for the impact of the number of GNN layers on EGNN-CD, we increase the number of GNN layers in each dataset respectively. The results are shown in Fig. 5, and the performance of EGNN-CD increases first and then decreases with the increase of the number of layers, and the overall performance does not fluctuate much. This suggests that EGNN-CD is not overfitting, which may be related to the dropout strategy.

## VI. CONCLUSION

This paper proposes an *End-to-end Graph Neural Networks-based Cognitive Diagnosis* (EGNN-CD) model. EGNN-CD consists of three main parts: knowledge concept network (KCN), GNN feature extraction (GNNFE) and cognitive ability prediction (CAP). First, KCN is aninfographic that synthesises physical information about students, exercises, and knowledge concepts to build CD-related interactions. Second, GNNFE designed a four-channel graph neural network to learn the higher-order features and individual features of KCN end-to-end. Finally, CAP combines high-level information to predict a student's ability. We fully demonstrate on three real-world datasets that the proposed model achieves higher accuracy and significance in various aspects, especially outperforming existing models in cognitive diagnosis. Furthermore, we also explore the impact of the GNN module on the performance of the EGNN-CD model, which provides a basis for future cognitive diagnosis work.